\documentclass[a4paper,11pt]{article}

\usepackage{fourier}
\usepackage{color}
 \usepackage{graphicx}
\usepackage{url}
\usepackage[affil-it]{authblk}
\usepackage{amsmath}
\usepackage{wrapfig}
\usepackage{subfigure}

\usepackage[T1]{fontenc}
\usepackage{times}

\usepackage{paralist}  

\title{A Framework for Pupil Tracking with Event Cameras}

\author[*,1]{Khadija Iddrisu}
\author[2]{Waseem Shariff}
\author[1]{Suzanne Little}

\affil[1]{Dublin City University, Ireland}
\affil[2]{University of Galway, Ireland}

\date{}

\begin{document}
\maketitle
\thispagestyle{empty}

\begin{abstract}
Saccades are extremely rapid movements of both eyes that occur simultaneously, typically observed when an individual shifts their focus from one object to another. These movements are among the swiftest produced by humans and possess the potential to achieve velocities greater than that of blinks. The peak angular speed of the eye during a saccade can reach as high as 700°/s in humans, especially during larger saccades that cover a visual angle of 25°. Previous research has demonstrated encouraging outcomes in comprehending neurological conditions through the study of saccades. A necessary step in saccade detection involves accurately identifying the precise location of the pupil within the eye, from which additional information such as gaze angles can be inferred. Conventional frame-based cameras often struggle with the high temporal precision necessary for tracking very fast movements, resulting in motion blur and latency issues. Event cameras, on the other hand, offer a promising alternative by recording changes in the visual scene asynchronously and providing high temporal resolution and low latency.  By bridging the gap between traditional computer vision and event-based vision, we present events as frames that can be readily utilized by standard deep learning algorithms. This approach harnesses YOLOv8, a state-of-the-art object detection technology, to process these frames for pupil tracking using the publicly accessible Ev-Eye dataset.
Experimental results demonstrate the framework's effectiveness, highlighting its potential applications in neuroscience, ophthalmology, and human-computer interaction.

\end{abstract}

\section{Introduction}

Saccades have been extensively studied in the context of various neurological and psychiatric disorders, such as schizophrenia, concussions, traumatic brain injuries (TBI), and Parkinson's disease~\cite{bittencourt2013saccadic}. Additionally, the relationship between saccades and cognition has shown strong correlation in identifying neurological disorders~\cite{yang2024exploring}. Assessing saccade changes during non-visual cognitive tasks may help to identify subtle brain changes due to aging and neuro-degenerative diseases. To make saccade analysis broadly useful for both diagnostic and research purposes, it is crucial to identify specific eye movement parameters. Key metrics, including velocity and duration, must be extracted from data that reflects a wide range of patient characteristics, such as differing eye shapes, and iris, hair, and skin pigmentation. However, to analyze saccades, the first step is to detect the pupil.

Automatic and non-invasive pupil tracking is a significant task in computer vision with applicability in areas such as Human-Computer Interactions (HCI), Virtual Reality and Extended Reality. This technology can contribute to greater understanding of psychological conditions by analysing the behaviour and tracking the movement of the pupil position over time. Recently, a few studies to improve pupil tracking have emerged~\cite{kang2019pupil,khan2020pupil,lee2020deep}. The majority of this research falls into two categories: remote pupil tracking and near-eye pupil tracking. Remote-eye pupil tracking involves pupil tracking from a distance from the eye which is applicable in areas such as driver monitoring while the latter involves pupil tracking with a setup in direct contact with the eyes (e.g., AR/VR). For effective pupil detection in non-clinical, real-world, environments, optimized sensors, such as event cameras, are essential.

Event cameras (ECs) have gained significant attention from researchers for pupil tracking due to their high sampling rate \cite{gallego2020event,kang2023exploring,shariff2024event,wang2024event}. ECs record changes in light intensity per pixel, offering unique features such as high dynamic range (HDR), high temporal resolution, and low latency. These characteristics open new opportunities for tasks like pupil tracking. The HDR capability of ECs allows them to capture and report changes occurring in very rapid motion, which often results in motion blur in traditional cameras.  Additionally, since ECs only capture changes in brightness levels, they protect user privacy by not collecting detailed iris data, as long as no reconstruction is done on the event stream~\cite{gallego2020event,shariff2024event}. Nonetheless, sensitive information might still be recoverable from event data through high resolution reconstruction~\cite{rebecq2019events}. Techniques such as  "event encryption" ~\cite{zhang2024event}, "Event-based person identification", "event scrambling"~\cite{ahmad2024event} among others have been proposed to enhance privacy  by encrypting the event data to prevent reconstruction.

Early studies on pupil tracking utilized computer vision and machine learning techniques, including pupil localization, eye center estimation, deep learning approaches, and support vector machines ~\cite{klaib2021eye}. Most of these studies relied on traditional frame-based camera systems and sometimes integrated near-infrared (NIR) pupil-tracking tasks. Inspired by recent advancements~\cite{rebecq2019events} which aims to bridge the gap between event cameras and modern computer vision, we propose using event-to-video conversion to generate event windows. These windows are then processed using YOLOv8, a state-of-the-art object detector, to detect and track the pupil accurately.

The contributions of this work are as follows:
\begin{compactenum}
    \item Integration of advanced computer vision techniques with the unique capabilities of event cameras by applying the YOLOv8 object detection model to event camera data. This combination enhances the accuracy and efficiency of pupil detection and tracking, leveraging the high temporal resolution and low latency of event cameras.
    
    \item Training of four different neural networks to process and analyze event camera data. This allows for a comprehensive evaluation of various models, identifying the most effective methods for pupil tracking and ensuring robust performance across different scenarios.

    \item A significant initial step towards the more complex goal of analyzing saccades. Focusing on the foundational task of accurate pupil detection and tracking using event cameras, the research sets the stage for future studies to investigate saccadic movements that are crucial for diagnosing and understanding various neurological and psychiatric disorders.

\end{compactenum}

\section{Related Work}

Despite the demonstrated potential of event-based pupil tracking, the body of research addressing this area remains limited. To date, only a small number of recent studies have undertaken this task. 

Kang et al~\cite{kang2023event} developed a large-scale event face training database through RGB-to-Event domain translation. They achieved this by using the StyleFlow algorithm to generate event-like images from RGB face images, leveraging existing annotations. Subsequently, the study evaluated a cross-modal learning-based pupil localization algorithm on real event camera images, captured with a DAVIS 346 camera, using a mixed dataset of RGB and event-like images. The algorithm, based on the RetinaFace algorithm and trained under this strategy, achieved a high accuracy in pupil localization, outperforming methods trained solely on RGB or event-like databases. In another study, the authors accumulated events over 33 ms intervals into frames~\cite{kang2023exploring}. Following this representation, they use cascaded Adaboost classifiers with multi-block local binary patterns (LBPs) to detect eye-nose region on which a Shape-Deformation Model (SDM) is applied for precise alignment, utilizing SIFT features and a Support Vector Machine (SVM) to maintain tracking across frames.  


Kagemoto et al~\cite{kagemoto2023event} presents the first purely event-based approach for pupil tracking with the bright and dark pupil effect elicited by alternately blinking two near-infrared illumination sources. The pupil center is detected by manually defining the region of interest and calculating the center of gravity of the more eventful polarity. This method processes events in real-time at 2000 Hz, significantly faster than traditional ellipse fitting methods, which operate at 30 Hz. Comparative experiments demonstrated that this approach accurately tracks eye movements with a high correlation to target movements and reduces gaze detection delay to 0.04 seconds compared to 1.18 seconds with ellipse fitting.

Finally, Swift-Eye~\cite{zhang2024swift} proposed a method to handle occlusion by leveraging Timelens, an event-based frame interpolation network, to transform low frame-rate video and asynchronous event streams into high-speed video up to 5000fps. Multi-scale spatial feature extraction is performed using a swin-transformer and feature pyramid network (FPN) to create hierarchical features for pupil estimation. Additionally, a rotated bounding box detects the pupil region by accommodating the ellipse's rotation in near-eye images. An occlusion-ratio-based approach adapts the detection method according to the level of occlusion, while invalid estimates are linearly interpolated. 

Our proposed approach diverges from Swift-Eye by not relying on information from RGB frames, which many event-based pupil tracking methods still utilize. Instead, we rely exclusively on events converted into frames in a manner that preserves the asynchronicity of events while mitigating issues such as undersampling. This approach aligns with existing event frame-based methods but aims to enhance the reliability and accuracy of pupil tracking without the need for additional RGB data.

\section{Methodology}

\subsection{Event Representation}
Deep learning algorithms such as Convolutional Neural Networks (CNNs) accept data inputs of fixed size such as images. Hence to perform pupil tracking with CNNs, there is a need to convert events into representations such as frames or voxel grids which can be readily used by these networks. In this study, we convert events into 2D frames by accumulating polarity pixel-wise.

Event cameras (ECs) possess an extremely high temporal resolution, theoretically capable of achieving up to 1 million frames per second (FPS) due to their microsecond ($1 \mu$) temporal precision~\cite{wang2019event}. This contrasts  with conventional frame-based cameras, which are typically limited to a maximum of about 100-145 FPS. We make use of the high temporal resolution by accumulating events over a 10 ms period to generate equivalent frames at 100 FPS. This approach allows us to significantly reduce the sampling time while preserving the temporal information inherent in the events. By doing so, we mitigate the risk of under-sampling and ensure that relevant information is maintained within the frames, all while preserving the asynchronous nature of the events. Additionally, this approach allows us to generate more frames for training (over 600 frames per event file) as compared to a few frames when accumulating based on a fixed size window. Sampling at 10 ms also aided in generation of frames captured during very rapid movements such as a half/almost full blink periods which may otherwise be lost in traditional imaging.

There are two methods commonly employed for 2D event accumulation; by accumulating in a fixed time window or fixed size window. The fixed time window approach conserves the temporal information of events, however for tasks such as pupil/eye tracking, it may lead to overlapping of eye features in time windows where there are many events generated due to motion. Therefore a fixed size approach is often employed for eye tracking applications, however due to the sparse nature of events, a larger window size is needed to make the pupil visible which results to low frame rate and subsequently can lead to under-sampling. 

In this study, we employ a fusion of both approaches such that, we sample a fixed duration window, i.e The total duration of events, from the minimum timestamp $\big(T_{\text{min}}\big)$ to the maximum timestamp $\big(T_{\text{max}}\big)$, is divided into frames of duration 10 milliseconds (ms) each. The number of frames $\big(N_{\text{frames}}\big)$ is determined as $(N_{\text{frames}} = \left\lceil \frac{T_{\text{max}} - T_{\text{min}}}{\text{duration\_ms}} \right\rceil)$. For each frame, events occurring within the time window $([T_{\text{start}}, T_{\text{end}}))$ are selected, where $(T_{\text{start}} = T_{\text{min}} + i \times \text{duration\_ms})$ and $(T_{\text{end}} = T_{\text{start}} + \text{duration\_ms})$. 

This process ensures that each frame represents a specific interval of time. To avoid creating frames with too few events, a threshold is applied. Frames are only generated if the number of events within the current time window exceeds the specified event threshold (set to 2000). This helps to filter out frames that might be empty or contain too little information.  The events within the selected time window are then processed to update the pixel values in the frame. For each event \((t, y, x, p)\), we check the event's coordinates and polarity. If the polarity \(p\) is +1, indicating an ON event, the corresponding pixel at \((y, x)\) is set to 255 (white). Conversely, if the polarity \(p\) is -1, indicating an OFF event, the pixel is set to 0 (black).

\subsection{Pupil Tracking with Yolo}
\begin{wrapfigure}{r}{0.5\textwidth}
  \vspace{-20pt}
  \begin{center}
    \includegraphics[width=0.5\textwidth, height=0.28\textwidth]{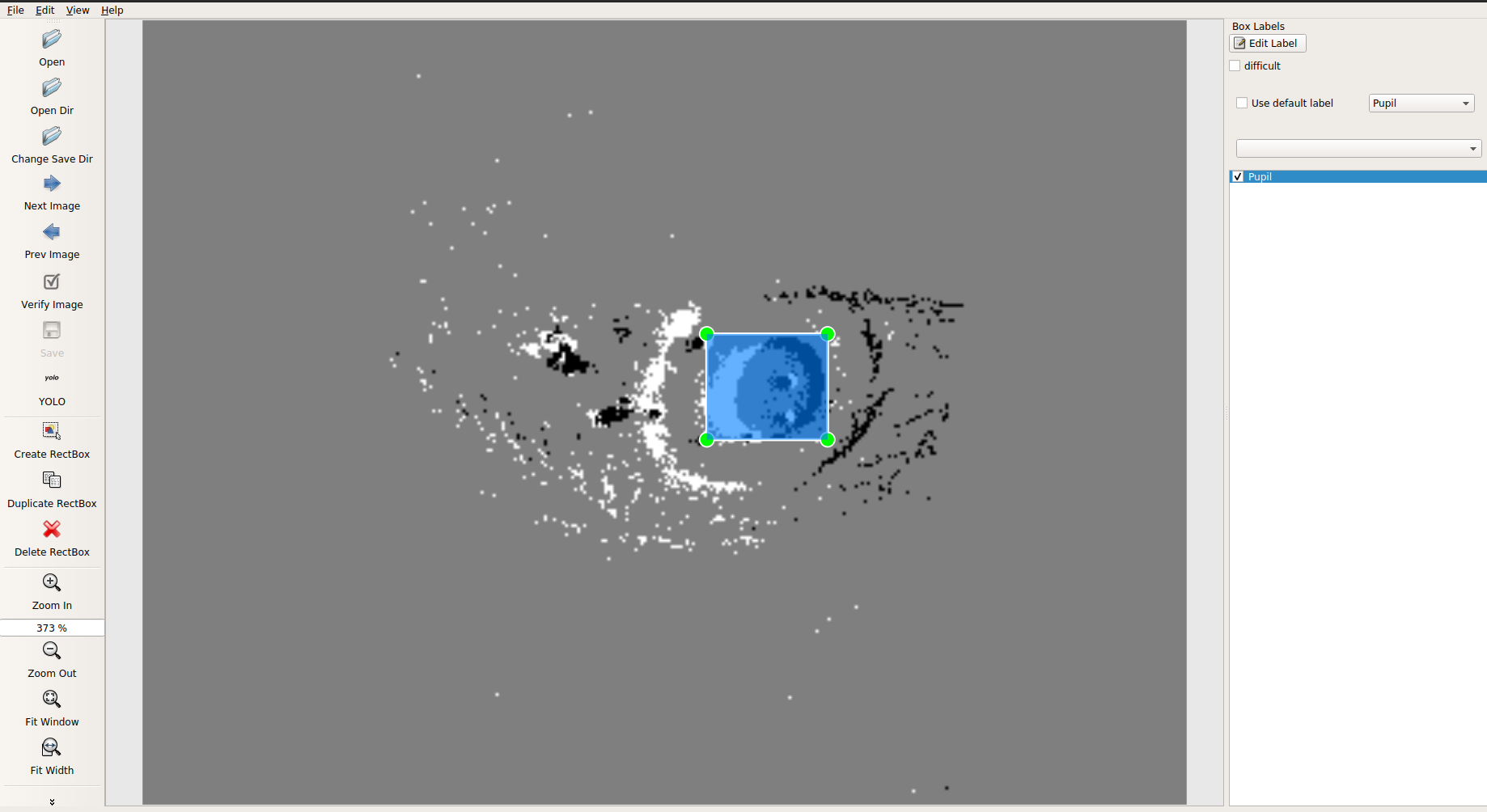}
\end{center}
\vspace{-20pt}
  \caption{Labelling process of event frames with LabelImg tool.}
  \label{Fig:LabelImg}
  \vspace{-0pt}
\end{wrapfigure}

\textbf{Data Preparation:} Following the method for event representations, we generated frames from the EV-Eye Dataset for training with YOLOv8. EV-Eye~\cite{zhao2024ev} consist of event data collected from 48 participants. Each participant participates in four sessions; fixation, saccades, and the last two session capturing smooth pursuit movement.  We utilise the raw unprocessed data and generate 20 random frames from each participant from both the left and right eye which allows us to generate a diverse range of frames from different eye movements for the resulting dataset. We select 38 subjects for training while reserving the remaining 10 for validation and testing. Subsequently, we generate 2400 frames from the training set and 400 each for the validation and test sets forming a total of 3200 frames. We then manually label these images with LabelImg~\cite{tzutalin2015labelimg} annotation tool with resulting bounding boxes in  YOLOv8 format for training. Fig \ref{Fig:LabelImg} illustrates the process of manual annotation.

\textbf{Training:} YOLOv8~\cite{Jocher_Chaurasia_Qiu_2023} offers enhanced accuracy and speed, making it suitable for real-time applications. YOLOv8 utilizes a novel architecture that builds upon the success of previous YOLO versions while introducing several key improvements. The backbone of YOLOv8 is a modified version of the CSPDarknet53 feature extractor, which employs cross-stage partial connections to enhance information flow between layers. Instead of the traditional Feature Pyramid Network (FPN), YOLOv8 utilizes a C2f module to combine high-level semantic features.  C2f represents two $3 \times 3$ convolutions with a residual connection which accepts outputs from the bottleneck and concatenates it. The bottleneck is the same as in  YOLOv5 with changes made to the size of the first convolutional layer from $1 \times 1$ to $3 \times 3$~\cite{sohan2024review}. The head of the network consists of multiple detection heads, each responsible for predicting bounding boxes, class probabilities, and objectness scores at different scales. YOLOv8 also incorporates a self-attention mechanism in the head to focus on relevant features and adjust their importance based on the task.  In our study, we utilised the event frames generated to train  YOLOv8. The dataset, comprising 3200 accumulated frames, was divided into a training, validation and test set with a ratio of 70:15:15 respectively. The training process was implemented in PyTorch and trained on an NVIDIA GeForce RTX 2080 Ti GPU. An AdamW optimizer was utilised in both models to achieve the highest performance with a learning rate of  $1 \times 10 ^{-3}$ and a weight decay of  $1 \times 10^{-3}$.

\section{Experiments and Results}

Table \ref{tab:mytab} presents the performance metrics of four different variants of the  YOLOv8 model (YOLOv8n,  YOLOv8s,  YOLOv8m, and  YOLOv8l) on the task of pupil tracking using the EV-EYE data. The models are evaluated based on mean average precision (mAP), precision, recall, F1 score, and the number of parameters (in millions).

In terms of mean average precision (mAP), which measures the accuracy and completeness of the model's predictions by calculating the average precision across all classes, YOLOv8n achieved the highest score of 0.981, indicating its superior performance in accurately identifying the pupil. The other models, YOLOv8s, YOLOv8m, and YOLOv8l, achieved slightly lower mAP scores, all around 0.975 to 0.976. Precision, which measures the accuracy of the positive predictions, was also highest for YOLOv8n at 0.965, signifying the lowest false positive rate among the models. The precision for the other models was slightly lower, with YOLOv8s at 0.950, YOLOv8m at 0.949, and YOLOv8l at 0.944.

Recall, which measures the ability to detect true positives, was highest for  YOLOv8l at 0.938.  True positive parameters are defined by two key thresholds: the Intersection over Union (IoU) threshold and the confidence threshold. The IoU threshold determines the minimum overlap required between a predicted bounding box and a ground truth bounding box for the prediction to be considered a true positive. An IoU threshold of 0.5 is used, meaning the predicted box must have at least 50\% overlap with the ground truth box to be counted as a true positive. The confidence threshold, on the other hand, defines the minimum confidence score a prediction must have to be considered a true positive. YOLOv8n had a recall of 0.919, which was slightly lower than the other models, ranging from 0.92 to 0.927.

The F1 score, which is the harmonic mean of precision and recall, was highest for  YOLOv8n and  YOLOv8l at 0.94.  YOLOv8s and  YOLOv8m had a slightly lower F1 score of 0.93. Regarding the number of parameters,  YOLOv8n had the fewest at 3.0 million, making it the most lightweight model. The number of parameters increased with model size:  YOLOv8s had 11.1 million,  YOLOv8m had 25.8 million, and  YOLOv8l had 43.6 million.

In summary, the  YOLOv8n model, despite having the fewest parameters, performed exceptionally well across all metrics, particularly excelling in mAP and precision. This makes it a highly efficient model for pupil tracking, balancing accuracy with computational efficiency. The larger models,  YOLOv8s,  YOLOv8m, and  YOLOv8l, showed strong performance as well, with slight variations in precision, recall, and F1 score, suggesting that while they may offer slight improvements in certain areas, they do so at the cost of increased computational complexity.

\begin{figure}[!ht]
    \centering
    \includegraphics[scale = 0.48]{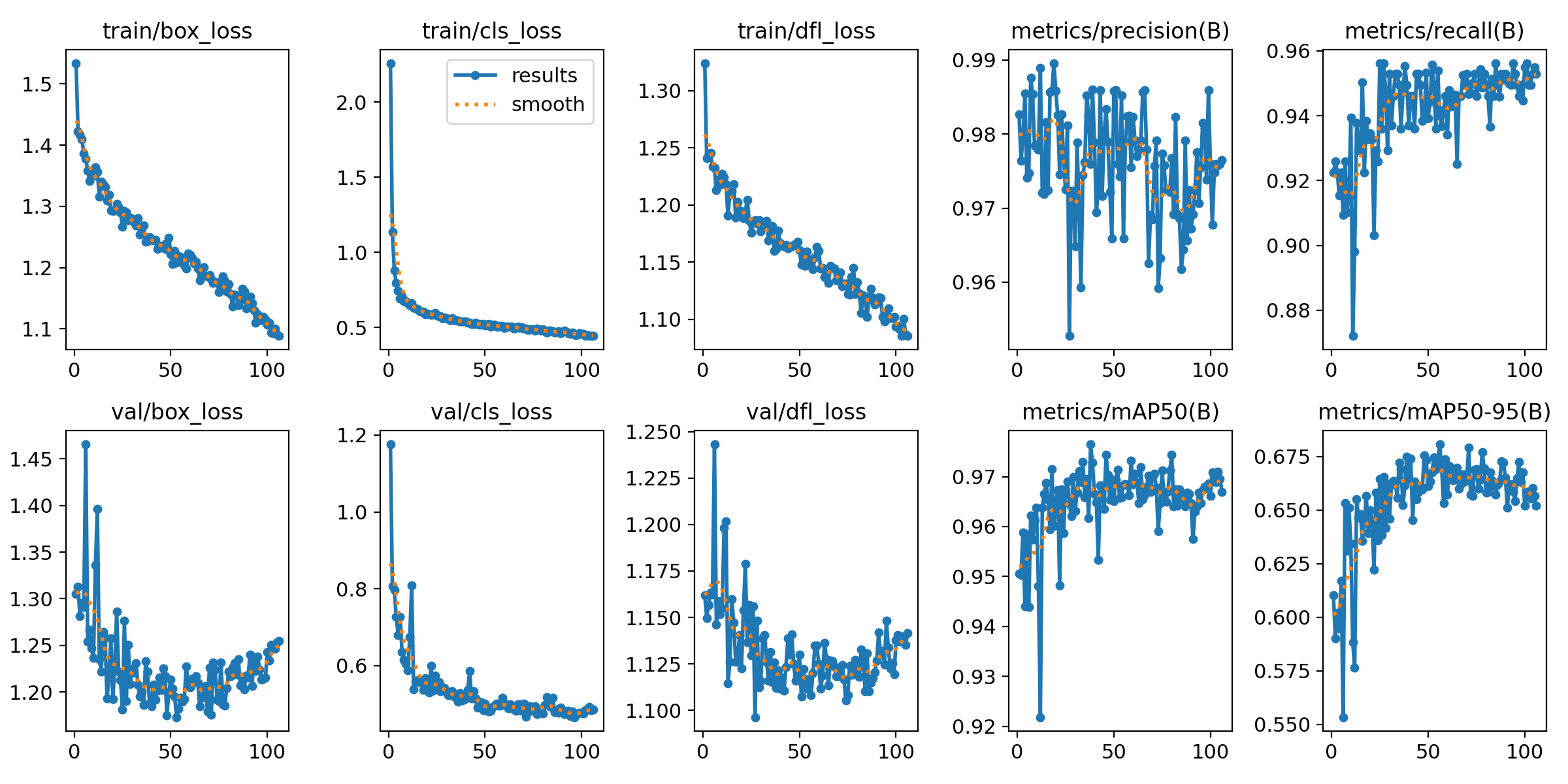}
    \caption{Training and Validation Losses for YOLOv8 on pupil tracking. The horizontal axes represent the number of epochs, while the vertical axes represent the value of each metric during training.}
    \label{fig:loss}
\end{figure}

\begin{table}[!h]
\begin{center}
\begin{tabular}{|c|c|c|c|c|r|}
\hline
YOLOv8 Model & Mean Average Precision &Precision & Recall & F1 Score & Parameters (M)\\
\hline
YOLOv8-n (nano)& \textbf{0.981} & \textbf{0.965} &    0.919 & \textbf{0.94} &\textbf{3.0}\\
\hline
YOLOv8-s (small)& 0.976 &0.950  &  0.920 & 0.93 & 11.1 \\
\hline
YOLOv8-m (medium)& 0.976 & 0.949   &  0.927 &0.93 & 25.8 \\
\hline
YOLOv8-l (large)& 0.975 & 0.944  & \textbf{ 0.938}  & 0.94 &43.6\\
\hline

\end{tabular}
\end{center}
\vspace{-20pt}
\caption{Results of 4 different  YOLOv8 variants trained using the EV-EYE data} \label{tab:mytab}
  \vspace{-10pt}
\end{table}
Figure \ref{fig:loss} shows the training and validation loss curves for YOLOv8-n on the pupil tracking task. These curves illustrate how the models' performance improved over the training period, indicating the effectiveness of the training process and the stability of the models. The upper rows present the training outcomes, whereas the lower rows show the results on the validation set.
It is observed that the loss overall decreases with more epochs indicating high accuracy. The first three columns depict the box loss, object loss, and classification loss of the improved YOLOv8 model. Additionally, it is observed that the validation losses exhibit greater noise and are marginally higher than the training losses, indicating some degree of overfitting. The precision, recall, and mAP metrics display more variability in the validation sets compared to the training sets. Therefore, although the model demonstrates good learning behavior, the fluctuations and slight inconsistencies in the validation metrics suggest a susceptibility to overfitting. Potential solutions to address this issue in future work include incorporating attention models and assembling YOLO variants.

\begin{figure}[!ht]
    \centering
    \includegraphics[scale = 0.9]{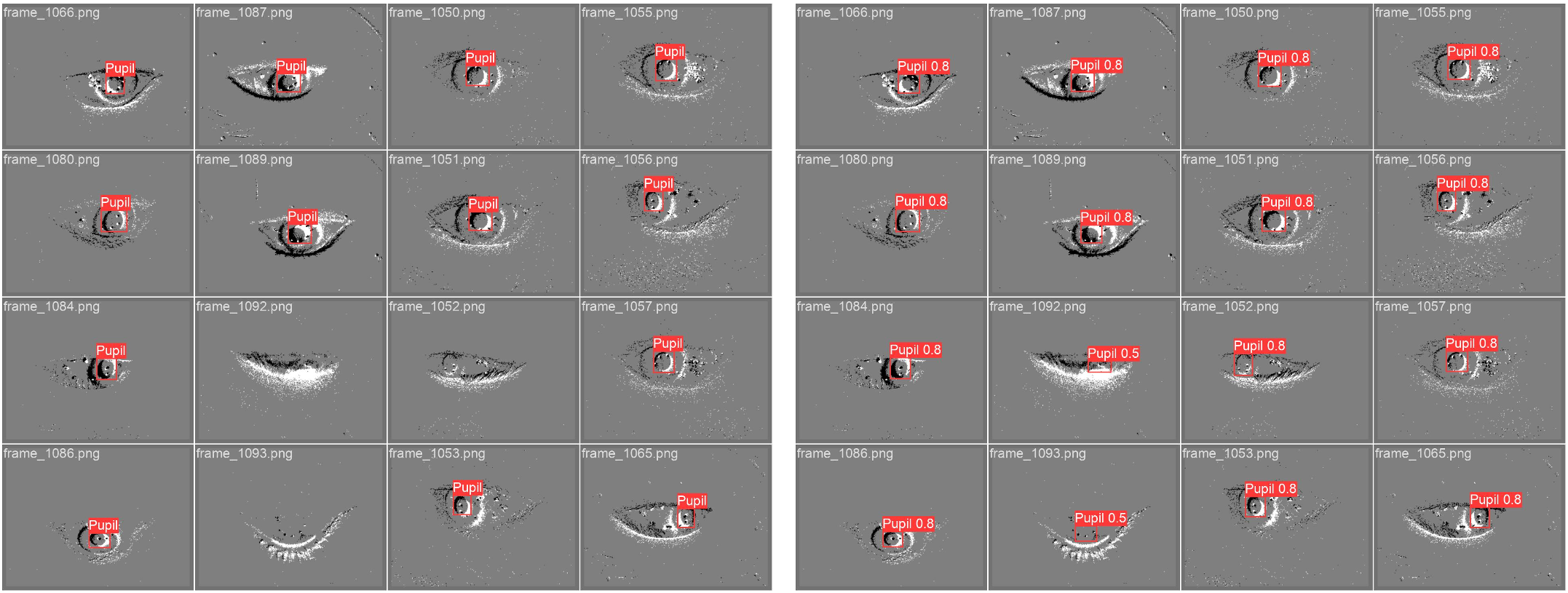}
    \caption{Qualitative results of the best performing proposed model (YOLOv8-n).The image on the left indicates ground truth labels while the image on the right indicates predicted labels along with confidence scores.}
    \label{fig:results}
\end{figure}

Figure \ref{fig:results} displays the qualitative results of pupil detection. As seen in the labels on the right hand side of the image, some images were not annotated to avoid inaccuracies in frames where the pupil were not fully visible. This happens during certain eye movements such as blinks when the events generated from movement from other eye parts is lesser than the pupil. However, the model demonstrates robust performance  by localizing the position of the pupil even when it is half visible. It is evident that the model is highly robust, maintaining pupil detection across various eye movements, even when the eye is nearly (frame\_1092) closed. Whether moving left or right(frame\_1052), the model consistently demonstrates robust performance.




\section{Limitations}
Our current model faces challenges when tested on remote datasets, which is not surprising given that we trained the models using a near-eye dataset. Although our focus is on applications in both near and remote eye sensing, we did not have a sufficient remote-eye dataset for training. As a result, the model's performance in identifying pupils in the presence of occlusions is less robust, and this may lead to significant errors. To address this issue, we plan to focus on acquiring and incorporating comprehensive remote eye datasets for further experiments, and we also intend to incorporate Region-of-Interest (ROI) techniques for directly extracting eye features to ensure robustness in all conditions and even in the presence of occlusions. This addition is expected to enhance the accuracy and robustness of our pupil tracking pipeline, paving the way for a saccade detection algorithm that can provide more reliable and precise outcomes. However, it is important to note that the model still performs fairly well in terms of pupil localization. We show this in the results presented in supplementary materials.

Additionally, while the YOLOv8n model performed exceptionally well in terms of mean average precision (mAP) and precision, its recall was slightly lower compared to the other models. Specifically, the YOLOv8n had a recall of 0.919, which is lower than the recall of 0.938 for YOLOv8l, and slightly below the recall values for YOLOv8s and YOLOv8m. This suggests that the YOLOv8n model may miss some true positives, which can be critical in applications where detecting every instance is essential.

\section{Conclusion}
In this study, we demonstrated pupil tracking in event cameras using a frame-based representation of events.  
By converting data from Event Cameras into a format understood by standard deep learning algorithms, we overcame issues such as under-sampling by generating events with duration of 10ms (100 fps) and highlighted the potential of the high frame rate offered by ECs. The presented results affirm the reliability and applicability of event camera imaging in pupil tracking. Our findings and  evaluation of various iterations of YOLOv8 highlights how deep learning algorithms can be adopted and transferred to dense representations such as event frames. ECs have demonstrated promising potential for object detection, especially in scenarios involving rapid motion and have the potential for the detection of high-speed changes. The use of ECs for pupil tracking has offered opportunities for high speed tracking in applications in XR, VR, Human Computer Interactions, e.t.c. 

Moving forward, the integration of ECs in pupil tracking presents exciting prospects for other areas involving analyzing eye movement. For instance,  saccade detection  which is essential for understanding human cognition and improving visual diagnostics of neurological conditions. The potential of such applications can lead to new avenues and breakthroughs in non-invasive diagnosis of neurological conditions such as Dementia and Parkinson's disease with further research and development required.

\section*{Acknowledgments}
This research was conducted with the financial support of Science Foundation Ireland under grant no. [12/RC/2289\_P2] at the Insight SFI Research Centre for Data Analytics, Dublin City University in collaboration with FotoNation (Tobii).

\appendix

\section{Appendix 1}
We evaluate the robustness of the proposed approach by testing on data from different variations of ECs and include the results in supplementary materials. 
\bibliographystyle{apalike}

\bibliography{imvip}

\end{document}